\documentclass[11pt,a4paper]{article}
\usepackage[hyperref]{emnlp2020}
\usepackage{times}
\usepackage{latexsym}

\usepackage{microtype}

\usepackage{amsmath}
\usepackage{amssymb}
\usepackage{graphicx}
\usepackage{enumitem}
\usepackage{booktabs}
\usepackage{mathtools}

\newcommand*{\Scale}[2][4]{\scalebox{#1}{$#2$}}%
\DeclareMathOperator*{\argmax}{argmax}

\usepackage{CJKutf8}

\aclfinalcopy 

\title{An Unsupervised Sentence Embedding Method by \\ Mutual Information Maximization}

\author{Yan Zhang\thanks{$^{*}$ Equally Contributed. This work was  done when Yan Zhang was an intern at DAMO Academy, Alibaba Group.}$^{~1,2}$~~~ Ruidan He$^{*}$$^{2}$~~~ Zuozhu Liu\thanks{~~Corresponding author.}$^{~3}$~~~ Kwan Hui Lim$^{1}$~~~Lidong Bing$^{2}$ \\
	$^{1}$Singapore University of Technology and Design \\
	$^{2}$DAMO Academy, Alibaba Group\\
	$^{3}$ZJU-UIUC Institue\\
	\texttt{yan\_zhang@mymail.sutd.edu.sg, ruidan.he@alibaba-inc.com}\\
	\texttt{zuozhuliu@intl.zju.edu.cn, kwanhui\_lim@sutd.edu.sg} \\
	\texttt{l.bing@alibaba-inc.com}}
  
\date{}

\begin{document}
\maketitle

\begin{abstract}


BERT is inefficient for sentence-pair tasks such as clustering or semantic search as it needs to evaluate combinatorially many sentence pairs which is very time-consuming. Sentence BERT (SBERT) attempted to solve this challenge by learning semantically meaningful representations of single sentences, such that similarity comparison can be easily accessed. However, SBERT is trained on corpus with high-quality labeled sentence pairs, which limits its application to tasks where labeled data is extremely scarce. In this paper, we propose a lightweight extension on top of BERT and a novel self-supervised learning objective based on mutual information maximization strategies to derive meaningful sentence embeddings in an unsupervised manner. 
Unlike SBERT, our method is not restricted by the availability of labeled data, such that it can be applied on different domain-specific corpus.
Experimental results show that the proposed method significantly outperforms other unsupervised sentence embedding baselines on common semantic textual similarity (STS) tasks and downstream supervised tasks. It also outperforms SBERT in a setting where in-domain labeled data is not available, and achieves performance competitive with supervised methods on various tasks. Our code is available at \url{https://github.com/yanzhangnlp/IS-BERT}.

\end{abstract}
\section{Introduction}

BERT-based pretrained language models \cite{devlin2019bert, liu2019roberta} have set new state-of-the-art performance on various downstream NLP tasks. However, they are inefficient for sentence-pair regression tasks such as clustering or semantic search because they need to evaluate combinatorially many sentence pairs during inference, which will result in a massive computational overhead. For example, finding the most similar pair in a collection of 10k sentences requires about 50 million ($10k \choose 2$)  inference computations with BERT, which requires about 65 hours on a V100 GPU \cite{reimers2019}. 

Much previous work attempted to address this problem by learning semantically meaningful representations for each sentence, such that similarity measures like cosine distance can be easily evaluated for sentence-pair regression tasks. The straightforward way to derive a fixed-size sentence embedding from BERT-based models is to average the token representations at the last layer or using the output of the [CLS] token. \citeauthor{reimers2019} (\citeyear{reimers2019}) showed that both approaches yield rather unsatisfactory sentence embeddings. They proposed a model, Sentence-BERT (SBERT), to further fine-tune BERT on natural language inference (NLI) tasks with labeled sentence pairs and achieved state-of-the-art performance on many semantic textual similarity tasks. However, such improvements are induced by high-quality supervision, 
and we find that their performance is degraded where labeled data of the target task is extremely scarce or the distribution of test set differs significantly from the NLI dataset used for training. 

Learning sentence representations in an unsupervised manner is a critical step to work with unlabeled or partially labeled dataset to address the aforementioned challenge \cite{Kiros2015,gan2017,hill2016,pagliardini2017unsupervised,Yang2018LearningST}. A common approach for unsupervised sentence representation learning is to leverage on self-supervision with large unlabeled corpus. For example, early methods explored various auto-encoders for sentence embedding \cite{Socher2011,hill2016}. Recent work such as skip-thought \cite{Kiros2015} and FastSent \cite{hill2016} assumed that a sentence is likely to have similar semantics to its context, and designed self-supervised objectives that encourage models to learn sentence representations by predicting contextual information. However, the performance of these models is far behind that of supervised learning ones on many tasks, which unveils an urgent need of better unsupervised sentence embedding methods. 
 
In this work, we propose a novel unsupervised sentence embedding model with light-weight feature extractor on top of BERT for sentence encoding, and train it with a novel self-supervised learning objective. Our model is not restricted by the availability of labeled data and can be applied to any domain of interest. Instead of simply averaging BERT token embeddings, we use convolutional neural network (CNN) layers with mean-over-time pooling that transform BERT token embeddings to a global sentence embedding \cite{kim2014convolutional}. Moreover, we propose a novel self-supervised learning objective that maximises the mutual information (MI) between the global sentence embedding and all its local contexts embeddings, inspired by recent advances on unsupervised representation learning for images and graphs \citep{hjelm2019,Velickovic2018DeepGI}. Our model is named Info-Sentence BERT (IS-BERT). In IS-BERT, the representation of a specific sentence is encouraged to encode all aspects of its local context information, using local contexts derived from other input sentences as negative examples for contrastive learning. This learning procedure encourages the encoder to capture the unique information that is shared across all local segments of the specific input sentence while different from other inputs, leading to more expressive and semantically meaningful sentence embeddings. 
 
We evaluate our method on two groups of tasks -- Semantic Textual Similarity (STS) and SentEval \cite{conneau2018}. Empirical results show that IS-BERT significantly outperforms other unsupervised baselines on STS and SentEval tasks. 
In addition, we show that IS-BERT substantially outperforms SBERT in a setting where task-specific labeled data is not available. This demonstrates that IS-BERT has the flexibility to be applied to new domains without label restriction. Finally, IS-BERT can achieve performance competitive with or even better than supervised learning methods in certain scenarios. 
 
\section{Related Work}\label{sec: related-work}

\subsection{Sentence Representation Learning}
Prior approaches for sentence embedding include two main categories: (1) unsupervised sentence embedding with unlabeled sentences, and (2) supervised learning with labeled sentences, while a few methods might leverage on both of them. 

\paragraph{Unsupervised Sentence Embedding.} There are two main directions to work with unlabeled corpus, according to whether the input sentences are ordered or not. In the scenario with unordered sentences, the input is usually a single sentence and models are designated to learn sentence representations base on the internal structures within each sentence,  such as recursive auto-encoders \cite{Socher2011}, denoising auto-encoders \cite{hill2016}, and the paragraph vector model \cite{Le2014}. Our model follows this setting as well but benefits from the model capacity of BERT and knowledge in large pretraining corpus. 

Methods working with ordered sentences utilize the distributional hypothesis which assumes that a sentence is likely to have similar semantics to its context. Under this assumption, they formulate generative or discriminative tasks that require the models to correctly predict the contextual information , such as skip-thought \cite{Kiros2015} and FastSent \cite{hill2016}, or to distinguish target sentences from contrastive ones \cite{Jernite17,logeswaran2018} for sentence embedding \cite{Jernite17,logeswaran2018}. These methods require ordered sentences or corpus with inter-sentential coherence for training, which limits their applications to domains with only short texts.

\paragraph{Supervised Sentence Embedding.} There have also been attempts to use labeled data for sentence embedding. \citeauthor{conneau2017} (\citeyear{conneau2017}) proposed the InferSent model that uses labeled data of the Stanford Natural Language Inference dataset (SNLI) \cite{bowman2015} and the Multi-Genre NLI dataset \cite{williams2018} to train a BiLSTM siamese network for sentence embedding. Universal Sentence Encoder \cite{cer2018} utilized supervised training with SNLI to augment the unsupervised training of a transformer network. SBERT \cite{reimers2019} also trained a siamese network on NLI to encode sentences, but it further benefits from the pretraining procedure of BERT. Though effective, those models could be problematic to port to new domains where high-quality labeled data is not available, or the text distribution is significantly different from the NLI dataset such that knowledge learned from NLI cannot be successfully transferred. Addressing this limitation requires unsupervised methods. 

\subsection{Representation Learning with MI}
Unsupervised representation learning with mutual information has a long history, such as the informax principle and ICA algorithms \cite{bell1995information,hyvarinen2000independent}. Theoretically, many generative models for representation learning based on reconstruction such as auto-encoders or GANs \cite{nowozin2016} are closely related to the idea of maximizing the MI between the model inputs and outputs. Despite the pivotal role in machine learning, MI is historically hard to compute, especially in high-dimensional and continuous settings such as neural networks. Recently, multiple estimators were proposed as lower bounds for mutual information estimation \cite{belghazi2018mine,oord2018}, which were demonstrated to be effective for unsupervised representation learning in various scenarios \cite{hjelm2019,ji2019,sun2019infograph,kong2020}. Our model is mainly inspired by the DIM model \cite{hjelm2019} for vision tasks, associated with a novel self-supervised learning objective to maximize the MI between the global sentence embedding and the representations of all its local contexts. Different from \cite{hjelm2019}, we mainly work with sequential sentence data with the pretrainted BERT model and further investigate the generalization ability of the learned representation across different domains. \citet{kong2020} also used MI with BERT, but their objective is for language modeling while our focus is on sentence representation learning. The corresponding downstream tasks are completely different as well. Pr
\section{Model}

\begin{figure*}
    \centering
    \includegraphics[scale=0.55]{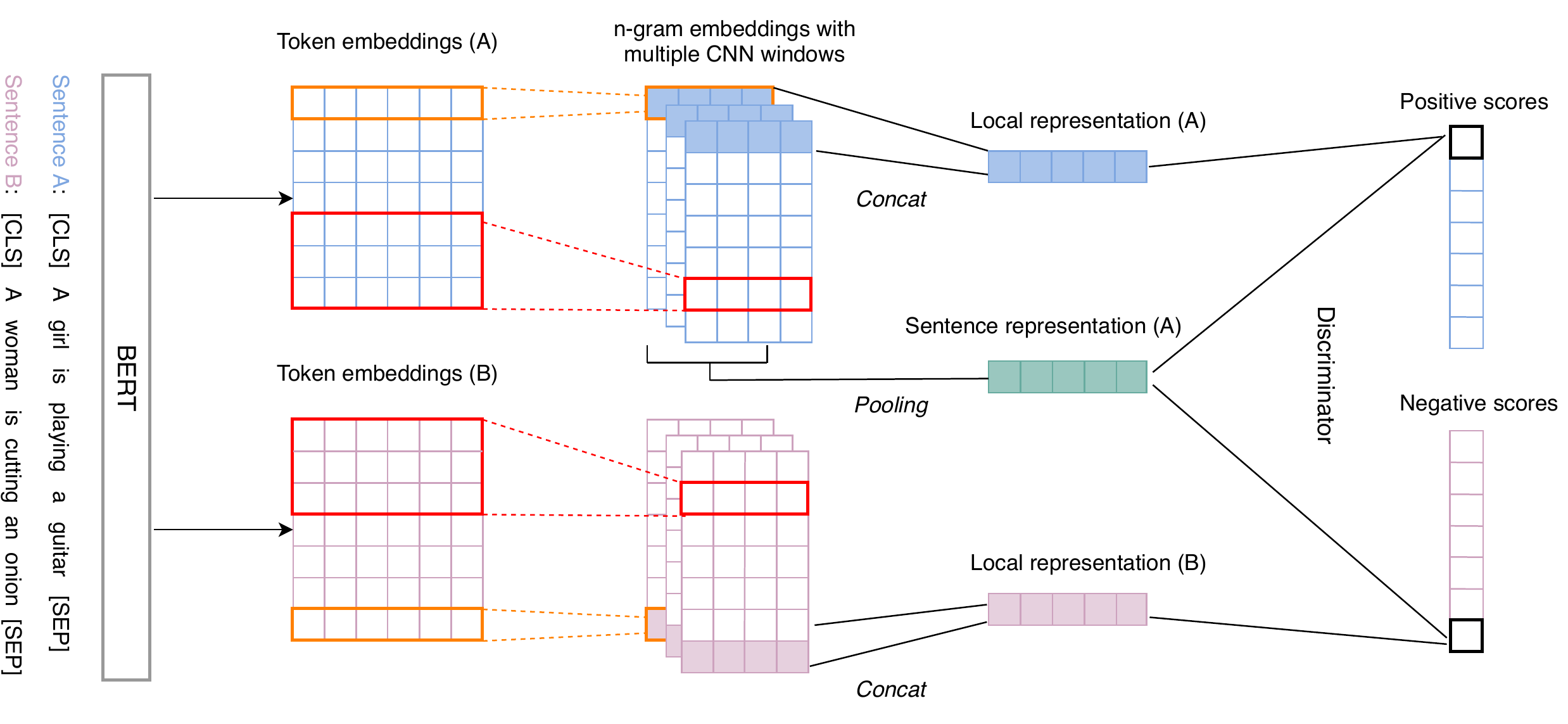}
    \vspace{-1mm}
    \caption{Model Architecture. Two sentences are encoded by BERT and multiple CNNs with different window sizes to get concatenated local n-gram token embeddings. A discriminator $T$ takes all pairs of \{sentence representation, token representation\} as input and decides whether they are from the same sentence. In this example, we treat sentence ``A'' as the positive sample and ``B" as negative, then n-gram embeddings of ``A'' will be summarized to a global sentence embedding via pooling. The discriminator produces scores for all token representations from both ``A" and ``B" to maximize the MI estimator in Eq.\ref{jsest}. 
    }
    \vspace{-5mm}
    \label{fig:layercomp}
\end{figure*}

In this section, we outline a general model, the Info-Sentence BERT (IS-BERT), for unsupervised sentence representation learning. We first give the problem formulation, then we present the details of our method and the corresponding neural network architecture. 

\subsection{Problem Formulation} 

Given a set of input sentences $\mathcal{X} = \{ \mathbf{x}_1, \mathbf{x}_2, ..., \mathbf{x}_n \}$, our goal is to learn a representation $\mathbf{y}_i \in R^{d}$ in $\mathcal{Y}$ for each sentence $\mathbf{x}_i$ in an unsupervised manner. For simplicity, we denote this process with a parameterized function $\mathcal{E}_{\Theta}: \mathcal{X} \longrightarrow \mathcal{Y}$, and denote the empirical distribution of the input set $\mathcal{X}$ as $\mathbb{P}$. 

We aim to acquire sentence representations by maximizing the mutual information between the sentence-level global representation and the token-level local representations. This idea was inspired by recent advances on unsupervised representation learning for images and graphs \citep{hjelm2019,sun2019infograph}. The motivation behind such learning strategy is to encourage sentence representations to encode multiple aspects shared by the local information of tokens such as n-gram contextual dependencies.



\subsection{Model Architecture}
Our model architecture is illustrated in Figure~\ref{fig:layercomp}. 
We first use BERT to encode an input sentence $\mathbf{x}$ to a length-$l$ sequence of token embeddings $\mathbf{h}_{1}, \mathbf{h}_{2}, ... , \mathbf{h}_{l}$. Then we apply 1-D convolutional neural network (CNN) layers with different window (kernel) sizes on top of these token embeddings to capture the n-gram local contextual dependencies of the input sentence. Formally, an n-gram embedding $\mathbf{c}_i$ generated by a CNN with window size $k$ is computed as
\begin{equation}
    \mathbf{c}_i = f(\mathbf{w} \cdot \mathbf{h}_{i:i+k-1}+ \mathbf{b} ),
\end{equation}
where $\mathbf{h}_{i:i+k-1}$ is the concatenation of the token embeddings within a window. $\mathbf{w}$ and $\mathbf{b}$ are learnable parameters of the CNN layer shared across all windows over the sequence, and $f$ is the ReLU activation. We use padding to keep the sequence length of outputs the same as inputs. 


To better capture contextual information with various ranges, we apply several CNNs with different window sizes (e.g. 1, 3, 5) to the input sentences. The final local representation of a token is the concatenation of its representations obtained with different window sizes, as shown in Figure~\ref{fig:layercomp}. We denote the length-$l$ local token representations sequence for a sentence $\mathbf{x}$ as $\mathcal{F}_\theta(\mathbf{x}) \coloneqq \{\mathcal{F}_\theta^{(i)} (\mathbf{x}) \in \mathbb{R}^{d}\}_{i=1}^{l}$, where $\mathcal{F}_\theta$ is the encoding function consisting of BERT and CNNs with trainable parameters $\theta$, and $i$ is the token index. The global sentence representation of $\mathbf{x}$ denoted as $\mathcal{E}_\theta(\mathbf{x}) \in \mathbb{R}^{d}$ is computed by applying a mean-over-time pooling layer on the token representations $\mathcal{F}_\theta (\mathbf{x})$. Both sentence and token representations are parameterized by $\theta$ as pooling does not  introduce additional parameters. The induction of the these representations is different from the previous sentence-BERT model \cite{reimers2019}. While \citet{reimers2019} simply used mean- or max-pooling strategies over the token representations from BERT outputs which can be regard as 1-gram embeddings, we use a set of parallel CNN layers with various window sizes to capture n-gram contextual dependencies. 
Both the sentence representation and token representations will be fed into a discriminator network to produce scores for MI estimation as presented in \ref{mim}. 

\subsection{MI Maximization Learning}
\label{mim}
The learning objective is to maximize the mutual information (MI) between the global sentence representation $\mathcal{E}_\theta(\mathbf{x})$ and each of its local token representation $\mathcal{F}_\theta^{(i)} (\mathbf{x})$. 
As MI estimation is generally intractable for continuous and high-dimensional random variables, we usually maximizing over lower bound estimators of MI, such as the Noise-Contrastive estimator \cite{gutmann2012noise} and Jensen-Shannon estimator \cite{nowozin2016, hjelm2019}. In this paper, we use the Jensen-Shannon estimator. Mathematically, the Jensen-Shannon estimator $\mathcal{I}_{\omega}^{JSD}(\mathcal{F}_\theta^{(i)}(\mathbf{x});\mathcal{E}_\theta(\mathbf{x}))$  is defined as 
\begin{equation}
\Scale[1]{
  \begin{aligned}
     \widehat{\mathcal{I}}_{\omega}^{JSD} & (\mathcal{F}_\theta^{(i)}(\mathbf{x});  \mathcal{E}_\theta(\mathbf{x})) \coloneqq \\
    & E_{\mathbb{P}}[-sp(-T_\omega(\mathcal{F}_\theta^{(i)}(\mathbf{x}),   \mathcal{E}_\theta(\mathbf{x})))]  \\
    & - E_{\mathbb{P} \times \tilde{\mathbb{P}}}[sp(T_\omega(\mathcal{F}_\theta^{(i)}(\mathbf{x}^{\prime}), \mathcal{E}_\theta(\mathbf{x})))],
  \end{aligned}
  }
  \label{jsest}
\end{equation}
where $T_\omega:\mathcal{F} \times \mathcal{E} \xrightarrow{} \mathbb{R}$
is a discriminator parameterized by a neural network with learnable parameters $\omega$. It takes all the pairs of a global sentence embedding and local token embeddings as input and generates corresponding scores to estimate $\widehat{\mathcal{I}}_{\omega}^{JSD}$, see Figure~\ref{fig:layercomp}. 
$\mathbf{x}^{\prime}$ is the negative sample drawn from distribution $\tilde{\mathbb{P}} = \mathbb{P}$, and $sp(z)=\log (1+e^z)$ is the softplus activation function. The end-goal learning objective over the whole dataset $\mathcal{X}$ is defined as:
\begin{equation}
\begin{aligned}
    {\omega}^{*}, \theta^{*} = &\argmax_{\omega, \theta} \frac{1}{|\mathcal{X}|} \Big( \\ &\displaystyle\sum_{\mathbf{x}\in \mathcal{X}} \displaystyle\sum_{i=1}^{l_{x}}\widehat{\mathcal{I}}_{\omega}^{JSD}  (\mathcal{F}_\theta^{(i)}(\mathbf{x});\mathcal{E}_\theta(\mathbf{x})) \Big),
\end{aligned}
\label{obj}
\end{equation}
where $|\mathcal{X}|$ is the size of the dataset, $l_x$ is the length of sentence $\mathbf{x}$, and ${\omega}^{*}, \theta^{*}$ denote the optimum. 

In Eq.~\ref{jsest}, $\mathcal{F}_\theta^{(i)}(\mathbf{x}^{\prime})$ corresponds to a local representation of the negative sample $\mathbf{x}^{\prime}$ drawn from $\widetilde{\mathbb{P}}=\mathbb{P}$. In practice, given a batch of sentences, we can treat each sentence and its local context representations as positive examples, and treat all the local context representations from other sentences in this batch as negative examples. Through maximizing $\widehat{\mathcal{I}}^{JSD}$, $\mathcal{E}_\theta (\mathbf{x})$ is encouraged to have high MI with its local context representations. 
This will push the encoder to capture the unique information that is shared across all local segments of the input sentence while different from other sentences, which leads to expressive sentence representation.


\section{Experiment}

\begin{table*}[t]
\centering
\begin{tabular}{l|ccccccc|c}
\toprule
\textbf{Model} &\textbf{STS12} &\textbf{STS13} &\textbf{STS14} &\textbf{STS15} &\textbf{STS16} &\textbf{STSb} &\textbf{SICK-R} & \textbf{Avg.}\\
\midrule
\midrule
\multicolumn{9}{l}{\emph{Using unlabeled data (unsupervised methods)}}\\
\midrule
Unigram-TFIDF$^{\dag}$ &- &- &58.00 &- &- &- &52.00 &-\\
SDAE$^{\dag}$ &- &- &12.00 &- &- &- &46.00 &-\\
ParagraphVec DBOW$^{\dag}$ &- &- &43.00 &- &- &- &42.00 &-\\
ParagraphVec DM$^{\dag}$ &- &- &44.00 &- &- &- &44.00 &-\\
SkipThought$^{\dag}$ &- &- &27.00 &- &- &- &57.00 &-\\
FastSent$^{\dag}$ &- &- &\bf{63.00} &- &- &- &61.00 &-\\
Avg. GloVe embeddings$^{\ddag}$ &55.14 &\bf{70.66} &59.73 &68.25 &63.66 &58.02 &53.76 &61.32\\
Avg. BERT embeddings$^{\ddag}$ &38.78 &57.98 &57.98 &63.15 &61.06 &46.35 &58.40 &54.81\\
BERT CLS-vector$^{\ddag}$ &20.16 &30.01 &20.09 &36.88 &38.08 &16.50 &42.63 &29.19\\
\hline 
\textbf{Ours: IS-BERT-NLI} &\bf{56.77} &69.24 &61.21 &\bf{75.23} &\bf{70.16} &\bf{69.21} &\bf{64.25} &\bf{66.58}\\
\bottomrule
\toprule
\multicolumn{9}{l}{\emph{Using labeled NLI data (supervised methods)}}\\
\hline 
InferSent - GloVe$^{\ddag}$ &52.86 &66.75 &62.15 &72.77 &66.87 &68.03 &65.65 &65.01\\
USE$^{\ddag}$ &64.49 &67.80 &64.61 &76.83 &73.18 &74.92 &76.69 &71.22\\
SBERT-NLI$^{\ddag}$ &70.97 &76.53 &73.19 &79.09 &74.30 &77.03 &72.91 &74.89\\ 
\bottomrule
\end{tabular}
\caption{Spearman rank correlation $\rho$ between the cosine similarity of sentence representations and the gold labels for various Semantic Textual Similarity (STS) tasks. $\rho * 100$ is reported in this paper. All BERT-based models use uncased-BERT-base as the transformer encoder. Results of baselines marked with $^{\dag}$ are extracted from \cite{hill2016} (with a different number of decimal places). Results of baselines marked with $^{\ddag}$ are extracted from \cite{reimers2019}. } \label{table:sts}
\end{table*}

Following previous works \cite{reimers2019,hill2016}, we conduct evaluation on two kinds of tasks:
\begin{itemize}[noitemsep,topsep=3pt]
    \item Unsupervised Semantic Textual Similarity (STS): These tasks measure a model's performance on sentence similarity prediction. The results are good indicators of effectiveness on unsupervised tasks such as clustering and semantic search. 
    \item Supervised downstream tasks: These tasks measure the effectiveness of sentence embeddings on downstream supervised tasks.
\end{itemize}

We consider two groups of baselines. The first group corresponds to models trained with unlabeled sentences. This includes the \textbf{unigram-TFIDF} mdoel, the \textbf{Paragraph Vector} model \cite{Le2014}, the Sequential Denoising Auto-Encoder (\textbf{SDAE})~\cite{hill2016}, the \textbf{Skipthought} \cite{Kiros2015} model and the \textbf{FastSent} \cite{hill2016} model,  all trained on the Toronto book corpus \cite{zhu2015} consisted of 70M sentences. We also consider representing sentence with the average of Glove embeddings, the average of the last layer representations of BERT, and the [CLS] embedding of BERT, respectively. 
The second group consists of models trained on labeled NLI data including \textbf{InferSent} \cite{conneau2017}, Universal Sentence Encoder (\textbf{USE}) \cite{cer2018}, and sentence BERT (\textbf{SBERT-NLI}) \cite{reimers2019}.
uncased-BERT-base is used for all BERT-related models including IS-BERT.


\subsection{Unsupervised Evaluations}
For STS tasks, we conduct evaluations on two types of datasets. \ref{sts_data} shows the results on seven STS benchmarks, which include texts from various domains and are commonly used for evaluating general-purpose sentence representations. \ref{afs_data} further shows the results on the challenging Argument Facet Similarity (AFS) dataset~\cite{misra2016}, which is more suitable for evaluating model's performance in domain-specific scenarios. 
For all methods compared in this subsection, cosine-similarity of the obtained sentence embeddings is used to compute their similarity, avoiding the time-consuming regression evaluation as with original BERT-based models. 

\subsubsection{Unsupervised STS} \label{sts_data}
\textbf{Experimental Details:}
We evaluate our model on the STS tasks 2012-2016 \cite{agirre2012, agirre2013, agirre2014, agirre2015, agirre2016}, the STS benchmark (STSb for short) \cite{cer2017}, and the SICK-Relatedness dataset \cite{marelli2014}. The corresponding datasets consist of sentence pairs with labels from 0 to 5 indicating the semantic relatedness.  As pointed out in \citet{reimers2016} that Pearson correlation is badly suited for STS,  Spearman's rank correlation between the cosine-similarity of the sentence embeddings and the gold labels is instead used as the evaluation metric. 

Following SBERT which was trained on the  combination of the SNLI \cite{bowman2015} and the Multi-Genre NLI (MultiNLI) \cite{williams2018} datasets with gold labels, we train IS-BERT on the same collection of sentences, but without using the label information. We denote our model in this setting as {\textbf{IS-BERT-NLI}}. SNLI contains 570,000 sentence pairs annotated with the labels \emph{contradiction}, \emph{entailment}, and \emph{neutral}. MultiNLI contains 430,000 sentence pairs which are from a wider range of genres of spoken and written texts. Note that IS-BERT-NLI was only trained on the 1 million pairs with the labels excluded.

We strictly follow the evaluation process of \citet{reimers2019} to make our results comparable to theirs. The development set of STSb is used for hyperparameter tuning. On all datasets, we apply three CNNs with window sizes 1, 3, and 5. Each CNN has 256 filters, making the final concatenated token representations of size 256*3. The learning rate is set to 1e-6 and the batch size is 32. 
\medskip

\noindent\textbf{Results: } Table \ref{table:sts} presents the results. Models are grouped into two sets by the nature of the data on which they were trained. We make the following observations. First, BERT out-of-the-box gives surely poor results on STS tasks. Both the [CLS] and averaging BERT embeddings perform worse than averaging GloVe embeddings. Second, all supervised methods outperform other unsupervised baselines, which suggests that the knowledge obtained from supervised learning on NLI can be well transfered to these STS tasks. This is also indicated in previous works \cite{hill2016,cer2018} that the dataset on which sentence embeddings are trained significantly impacts their performance on STS benchmarks and they found NLI datasets are particularly useful. 

On average, IS-BERT-NLI significantly outperforms other unsupervised baselines. It even outperforms InferSent trained on labeled SNLI and MultiNLI datasets in 5 out of 7 tasks. This demonstrates the effectiveness of the proposed training strategy. USE and SBERT are the top two performers. 
As expected, IS-BERT-NLI is in general inferior to these two supervised baselines because they are trained with the particularly useful labeled NLI data as well as large unlabeled data, but IS-BERT-NLI also achieves performance comparable to them in certain scenarios, e.g., STS13 and STS15, even it was only trained on the NLI unlabeled data. 



\subsubsection{Argument Facet Similarity} \label{afs_data}

We have shown in Section \ref{sts_data} that the proposed model substantially outperforms other unsupervised methods. 
However, the STS benchmarks in Section \ref{sts_data} are not domain or task specific, and it has been shown that they favor the supervised methods trained on NLI more \cite{hill2016,cer2018}. 
In this subsection, we further conduct evaluation on an Argument Facet Similarity (AFS) dataset \cite{misra2016} which is more task-specific. Models are compared in a setting without task or domain-specific labeled data. In this setting, SBERT needs to be trained on NLI and transferred to the AFS dataset for evaluation. Since IS-BERT does not require labeled data, it can be directly trained on the task-specific raw texts. We denote our model in this setting as {\textbf{IS-BERT-AFS}}.
\medskip

\noindent\textbf{Experimental Details:} The AFS corpus annotated 6,000 sentential argument pairs from social media dialogs on three controversial topics: \emph{gun control}, \emph{gay marriage}, and \emph{death penalty}. Each argument pair was annotated on a scale from 0 (different) to 5 (equivalent). To be considered similar, argument must not only make similar claims, but also provide a similar reasoning. In addition, the lexical gap between the sentences in AFS is much larger, making it a more challenging task compared to STS tasks. The proposed IS-BERT-AFS is trained on sentences from all three domains. It uses CNNs with window size set to 3, 5, and 7, as the average sentence length is longer in AFS.  Other hyperparameters are the same as in Section \ref{sts_data}.
\medskip

\noindent\textbf{Results: } Table \ref{table:afs} presents the results. We also provide the Pearson correlation $r$ to make the results comparable to \cite{reimers2019}. The models in the top group are trained without task-specific labeled data. IS-BERT-AFS clearly outperforms other models in this setting. One major finding is that SBERT-NLI and InferSent, the models trained on the labeled NLI data, perform the worst on this task. We believe this is due to the fact that the NLI corpus significantly differs from the AFS dataset. An improper training set could lead to extremely bad performance in the unsupervised transfer learning setting,  which supports our claim that supervised sentence embedding methods are problematic to port to new domains when the distribution of target data (i.e. AFS) differs significantly from the pretraining one (i.e. NLI). 

We also show results of BERT and SBERT in another two settings when trained with task-specific labeled data as in \cite{reimers2019}. When trained on all topics (10-fold cross-validation), both BERT and SBERT easily achieve scores above 70, while we observe large performance drop when they are trained in a cross-topic setting (i.e. train on two topics of AFS and evaluate on the third topic). The even larger performance drop of SBERT when trained on NLI ($\rho$ from 74.13 to 15.84) again demonstrates that the domain-relatedness between the training set and the target test set has a huge impact on supervised sentence embedding learning, as a result, the supervised methods are problematic to be applied to downstream tasks of domains without labeled data.

\subsection{Supervised Evaluations}
\subsubsection{SentEval}

\textbf{Experimental Details: } Here we evaluate the sentence representations in IS-BERT on a set of supervised tasks. 
Following \citet{reimers2019}, we use a set of classification tasks that covers various types of sentence classification, including sentiment analysis (\textbf{CR} \cite{Hu2004}, \textbf{MR} \cite{pang2005} and \textbf{SST} \cite{socher2013}), question-type classification (\textbf{TREC} \cite{li2002}), subjectivity classification (\textbf{SUBJ} \cite{pang2004}), opinion polarity classification (\textbf{MPQA} \cite{Wiebe2005}) and paraphrase identification (\textbf{MRPC} \cite{dolan2004}).

Since these tasks are more domain-specific, we train IS-BERT on each of the task-specific dataset (without label) to produce sentence embeddings, which are then used for training downstream classifiers. We denote this setting as \textbf{IS-BERT-task}.
SentEval \cite{conneau2018} toolkit is used to automate the evaluation process. It takes sentence embeddings as fixed input features to a logistic regression classifier, which is trained in a 10-fold cross-validation setup and the prediction accuracy is computed for the test-fold. 
\medskip

\begin{table}[t]
\centering
\begin{tabular}{l|cc}
\toprule
\textbf{Model}&$r$&$\rho$\\
\midrule
\midrule
\multicolumn{3}{l}{\emph{Without task-specific labeled data}}\\
\midrule
Unigram-TFIDF &46.77 &42.95\\
InferSent-GloVe &27.08 &26.63\\
Avg. GloVe embeddings &32.40 &34.00\\
Avg. BERT embeddings &35.39 &35.07\\
SBERT-NLI &16.27 &15.84\\
\hline
\textbf{Ours: IS-BERT-AFS} &\bf{49.14} &\bf{45.25}\\
\bottomrule
\toprule
\multicolumn{3}{l}{\emph{Supervised: 10-fold cross-validation}}\\
\hline
BERT-AFS &77.20 &74.84\\
SBERT-AFS &76.57 &74.13\\
\bottomrule
\toprule
\multicolumn{3}{l}{\emph{Supervised: cross-topic evaluation}}\\
\hline
BERT-AFS &58.49 &57.23\\
SBERT-AFS &52.34 &50.65\\
\bottomrule
\end{tabular}
\caption{Average Pearson correlation $r$ and average Spearman's rank correlation $\rho$ over three topics on the Argument Facet Similarity (AFS) corpus. Results of baselines are extracted from \cite{reimers2019, reimers2019b}} \label{table:afs}
\end{table}

\begin{table*}[t]
\centering
\scalebox{0.95}{
\begin{tabular}{l|ccccccc|c}
\toprule
\textbf{Model} &\textbf{MR} &\textbf{CR} &\textbf{SUBJ} &\textbf{MPQA} &\textbf{SST} &\textbf{TREC} &\textbf{MRPC} & \textbf{Avg.}\\
\midrule
\midrule
\multicolumn{9}{l}{\emph{Using unlabeled data (unsupervised methods)}}\\
\midrule
Unigram-TFIDF$^{\dag}$ &73.7 &79.2 &90.3 &82.4 &- &85.0 &73.6 &-\\
SDAE$^{\dag}$ &74.6 &78.0 &90.8 &86.9 &- &78.4 &73.7 &-\\
ParagraphVec DBOW$^{\dag}$ &60.2 &66.9 &76.3 &70.7 &- &59.4 &72.9 &-\\
SkipThought$^{\dag}$ &76.5 &80.1 &93.6 &87.1 &82.0 &92.2 &73.0 &83.50\\
FastSent$^{\dag}$ &70.8 &78.4 &88.7 &80.6 &- &76.8 &72.2 &-\\
Avg. GloVe embeddings$^{\ddag}$ &77.25 &78.30 &91.17 &87.85 &80.18 &83.0 &72.87 &81.52\\
Avg. BERT embeddings$^{\ddag}$ &78.66 &86.25 &94.37 &88.66 &84.40 &\bf{92.8} &69.54 &84.94\\
BERT CLS-vector$^{\ddag}$ &78.68 &84.85 &94.21 &88.23 &84.13 &91.4 &71.13 &84.66\\
\hline 
\textbf{Ours: IS-BERT-task} &\bf{81.09} &\bf{87.18} &\bf{94.96} &\bf{88.75} &\bf{85.96} &88.64 &\bf{74.24} &\bf{85.91}\\
\bottomrule
\toprule
\multicolumn{9}{l}{\emph{Using labeled NLI data (supervised methods)}}\\
\hline 
InferSent - GloVe$^{\ddag}$ &81.57 &86.54 &92.50 &90.38 &84.18 &88.2 &75.77 &85.59\\
USE$^{\ddag}$ &80.09 &85.19 &93.98 &86.70 &86.38 &93.2 &70.14 &85.10\\
SBERT-NLI$^{\ddag}$ &83.64 &89.43 &94.39 &89.86 &88.96 &89.6 &76.00 &87.41\\ 
\bottomrule
\end{tabular}}
\caption{Evaluation accuracy using the SentEval toolkit. Scores are based on a 10-fold cross-validation. Results of baselines marked with $^{\dag}$ are extracted from \cite{hill2016} (with a different number of decimal places). Results of baselines marked with $^{\ddag}$ are extracted from \cite{reimers2019}.} \label{table:senteval}
\end{table*}

\noindent\textbf{Results: } Table \ref{table:senteval} presents the results. Overall, supervised methods outperform unsupervised baselines. This indicates that pretraining sentence encoder with high-quality labeled data such as NLI is helpful in a supervised transfer learning setting. Note that in this task, SentEval fits a logistic regression classifier to the sentence embeddings with labels of the downstream tasks. Thus, the models that achieve good results on this task do not necessarily work well on unsupervised tasks such as clustering. As shown in Section \ref{afs_data}, training on NLI could lead to extremely bad performance on downstream unsupervised tasks when the domain data significantly differs from NLI.

IS-BERT-task is able to outperform other unsupervised baselines on 6 out of 7 tasks, and it is on par with InferSent and USE which are strong supervised baselines trained on NLI task. This demonstrates the effectiveness of the proposed model in learning domain-specific sentence embeddings. 

\subsubsection{Supervised STS}
\begin{table}[t]
\centering
\scalebox{0.9}{
\begin{tabular}{l|c}
\toprule
\textbf{Model}& {$\rho$}\\
\midrule
BERT-STSb &84.30 $\pm$ 0.76 \\
SBERT-STSb &84.67 $\pm$ 0.19 \\
\textbf{Ours: IS-BERT-STSb (ft)} & 74.25 $\pm$  0.94 \\
\textbf{Ours: IS-BERT-STSb (ssl + ft)} & \bf{84.84} $\pm$ 0.43\\
\bottomrule
\end{tabular}}
\caption{Spearman's rank correlation $\rho$ on the STSb test set. Results of baselines are extracted from \cite{reimers2019}. All systems are
trained with 10 random seeds to counter variances \cite{reimers2019}.}
\end{table}

\textbf{Experimental Details:} Following \citet{reimers2019}, we use the STSb \cite{cer2017} dataset to evaluate models' performance on the supervised STS task. This dataset includes 8,628 sentence pairs from the three categories \emph{captions}, \emph{news}, and \emph{forums}. It is divided into train (5,749), dev (1,500) and test (1,379). 

We compare IS-BERT to the state-of-the-art BERT and SBERT methods on this task. BERT is trained with a regression head on the training set with both sentences passed to the network (BERT-STSb). SBERT is trained on the training set by encoding each sentence separately and using a regression objective function. 

We experiment with two setups: 1) 
Without self-supervised learning with the max-MI objective in Eq.\ref{obj}, IS-BERT is directly used for encoding each sentence and fine-tuned on the training set with a regression objective. We denote this setting as \textbf{IS-BERT-STSb (ft)}. 2) IS-BERT is first trained on the training set without label using the self-supervised learning objective. Then, it is fine-tuned on the labeled data with a regression objective. We denote this setting as \textbf{IS-BERT-STSb (ssl+ft)}. 
At the prediction time, cosine similarity is computed between each pair of sentences. 
\medskip

\noindent\textbf{Results: } The results are depicted in Table 4. BERT and SBERT performs similarly on this task. IS-BERT-STSb (ssl+ft) outperforms both baselines. Another interesting finding is that when directly fine-tuning IS-BERT on the labeled data, it performs much worse than SBERT. The only difference between them is that IS-BERT-STSb(ft) uses CNN layers with mean pooling to obtain sentence embeddings while SBERT simply uses a pooling layer to do so. This suggests that a more complex sentence encoder does not 
automatically lead to better sentence embeddings. However, when comparing IS-BERT-STSb(ft) with IS-BERT-STSb(ssl+ft), we observe that adding self-supervised learning before fine-tuning leads to more than 10\% performance improvements. This indicates that the our self-supervise learning method can also be used as an effective domain-adaptation approach before fine-tuning the network.

\section{Conclusions}

In this paper, we proposed the IS-BERT model for unsupervised sentence representation learning with a novel MI maximization objective. IS-BERT outperforms all unsupervised sentence embedding baselines on various tasks and is competitive with supervised sentence embedding methods in certain scenarios. In addition, we show that sentence BERT (SBERT), the state-of-the-art supervised method, is problematic to apply to certain unsupervised tasks when the target domain significantly differs from the dataset it was trained on. IS-BERT achieves substantially better results in this scenario as it has the flexibility to be trained on the task-specific corpus without label restriction. In the future, we want to explore semi-supervised methods for sentence embedding and its transferability across domains. 

\bibliography{emnlp2020}

\begin{thebibliography}{47}
\expandafter\ifx\csname natexlab\endcsname\relax\def\natexlab#1{#1}\fi

\bibitem[{Agirre et~al.(2015)Agirre, Banea, Cardie, Cer, Diab, Gonzalez-Agirre,
  Guo, Lopez-Gazpio, Maritxalar, Mihalcea, Rigau, Uria, and Wiebe}]{agirre2015}
Eneko Agirre, Carmen Banea, Claire Cardie, Daniel Cer, Mona Diab, Aitor
  Gonzalez-Agirre, Weiwei Guo, Iñigo Lopez-Gazpio, Montse Maritxalar, Rada
  Mihalcea, German Rigau, Larraitz Uria, and Janyce Wiebe. 2015.
\newblock Semeval-2015 task 2: Semantic textual similarity, english, spanish
  and pilot on interpretability.
\newblock In \emph{Proc. of SemEval@ACL}.

\bibitem[{Agirre et~al.(2014)Agirre, Banea, Cardie, Cer, Diab, Gonzalez-Agirre,
  Guo, Mihalcea, Rigau, and Wiebe}]{agirre2014}
Eneko Agirre, Carmen Banea, Claire Cardie, Daniel Cer, Mona Diab, Aitor
  Gonzalez-Agirre, Weiwei Guo, Rada Mihalcea, German Rigau, and Janyce Wiebe.
  2014.
\newblock Semeval-2014 task 10: Multilingual semantic textual similarity.
\newblock In \emph{Proc. of SemEval@ACL}.

\bibitem[{Agirre et~al.(2016)Agirre, Banea, Cer, Diab, Gonzalez-Agirre,
  Mihalcea, Rigau, and Wiebe}]{agirre2016}
Eneko Agirre, Carmen Banea, Daniel Cer, Mona Diab, Aitor Gonzalez-Agirre, Rada
  Mihalcea, German Rigau, and Janyce Wiebe. 2016.
\newblock Semeval-2016 task 1: Semantic textual similarity, monolingual and
  cross-lingual evaluation.
\newblock In \emph{Proc. of SemEval@ACL}.

\bibitem[{Agirre et~al.(2012)Agirre, Cer, Diab, and
  Gonzalez-Agirre}]{agirre2012}
Eneko Agirre, Daniel Cer, Mona Diab, and Aitor Gonzalez-Agirre. 2012.
\newblock Semeval-2012 task 6: A pilot on semantic textual similarity.
\newblock In \emph{SemEval@ACL}.

\bibitem[{Agirre et~al.(2013)Agirre, Cer, Diab, Gonzalez-Agirre, and
  Guo}]{agirre2013}
Eneko Agirre, Daniel Cer, Mona Diab, Aitor Gonzalez-Agirre, and Weiwei Guo.
  2013.
\newblock *sem 2013 shared task: Semantic textual similarity.
\newblock In \emph{The Second Joint Conference on Lexical and Computational
  Semantics}.

\bibitem[{Belghazi et~al.(2018)Belghazi, Baratin, Rajeswar, Ozair, Bengio,
  Courville, and Hjelm}]{belghazi2018mine}
Mohamed~Ishmael Belghazi, Aristide Baratin, Sai Rajeswar, Sherjil Ozair, Yoshua
  Bengio, Aaron Courville, and R~Devon Hjelm. 2018.
\newblock Mine: mutual information neural estimation.

\bibitem[{Bell and Sejnowski(1995)}]{bell1995information}
Anthony~J Bell and Terrence~J Sejnowski. 1995.
\newblock An information-maximization approach to blind separation and blind
  deconvolution.
\newblock \emph{Neural computation}, 7(6):1129--1159.

\bibitem[{Bowman et~al.(2015)Bowman, Angeli, Potts, and Manning}]{bowman2015}
Samuel~R. Bowman, Gabor Angeli, Christopher Potts, and Christopher~D. Manning.
  2015.
\newblock A large annotated corpus for learning natural language inference.
\newblock In \emph{Proc. of EMNLP}.

\bibitem[{Cer et~al.(2017)Cer, Diab, Agirre, Lopez-Gazpio, and
  Specia}]{cer2017}
Daniel Cer, Mona Diab, Eneko Agirre, Iñigo Lopez-Gazpio, and Lucia Specia.
  2017.
\newblock Semeval-2017 task 1: Semantic textual similarity multilingual and
  cross-lingual focused evaluation.
\newblock In \emph{Proc. of SemEval@ACL}.

\bibitem[{Cer et~al.(2018)Cer, Yang, Kong, Hua, Limtiaco, St.~John, Constant,
  Guajardo-Cespedes, Yuan, Tar, Strope, and Kurzweil}]{cer2018}
Daniel Cer, Yinfei Yang, Sheng-yi Kong, Nan Hua, Nicole Limtiaco, Rhomni
  St.~John, Noah Constant, Mario Guajardo-Cespedes, Steve Yuan, Chris Tar,
  Brian Strope, and Ray Kurzweil. 2018.
\newblock Universal sentence encoder for {E}nglish.
\newblock In \emph{Proc. of EMNLP}.

\bibitem[{Conneau and Kiela(2018)}]{conneau2018}
Alexis Conneau and Douwe Kiela. 2018.
\newblock {S}ent{E}val: An evaluation toolkit for universal sentence
  representations.
\newblock In \emph{Proc. of LREC}.

\bibitem[{Conneau et~al.(2017)Conneau, Kiela, Schwenk, Barrault, and
  Bordes}]{conneau2017}
Alexis Conneau, Douwe Kiela, Holger Schwenk, Lo{\"\i}c Barrault, and Antoine
  Bordes. 2017.
\newblock Supervised learning of universal sentence representations from
  natural language inference data.
\newblock In \emph{Proc. of EMNLP}.

\bibitem[{Devlin et~al.(2019)Devlin, Chang, Lee, and
  Toutanova}]{devlin2019bert}
Jacob Devlin, Ming-Wei Chang, Kenton Lee, and Kristina Toutanova. 2019.
\newblock {BERT}: Pre-training of deep bidirectional transformers for language
  understanding.
\newblock In \emph{Proc. of NAACL-HLT}.

\bibitem[{Dolan et~al.(2004)Dolan, Quirk, and Brockett}]{dolan2004}
Bill Dolan, Chris Quirk, and Chris Brockett. 2004.
\newblock Unsupervised construction of large paraphrase corpora: Exploiting
  massively parallel news sources.
\newblock In \emph{Proc. of COLING}.

\bibitem[{Gan et~al.(2017)Gan, Pu, Henao, Li, He, and Carin}]{gan2017}
Zhe Gan, Yunchen Pu, Ricardo Henao, Chunyuan Li, Xiaodong He, and Lawrence
  Carin. 2017.
\newblock Learning generic sentence representations using convolutional neural
  networks.
\newblock In \emph{Proc. of EMNLP}.

\bibitem[{Gutmann and Hyv{\"a}rinen(2012)}]{gutmann2012noise}
Michael~U Gutmann and Aapo Hyv{\"a}rinen. 2012.
\newblock Noise-contrastive estimation of unnormalized statistical models, with
  applications to natural image statistics.
\newblock \emph{Journal of Machine Learning Research}, 13(Feb):307--361.

\bibitem[{Hill et~al.(2016)Hill, Cho, and Korhonen}]{hill2016}
Felix Hill, Kyunghyun Cho, and Anna Korhonen. 2016.
\newblock Learning distributed representations of sentences from unlabelled
  data.
\newblock In \emph{Proc. of NAACL-HLT}.

\bibitem[{Hjelm et~al.(2019)Hjelm, Fedorov, Lavoie-Marchildon, Grewal, Bachman,
  Trischler, and Bengio}]{hjelm2019}
Devon Hjelm, Alex Fedorov, Samuel Lavoie-Marchildon, Karan Grewal, Philip
  Bachman, Adam Trischler, and Yoshua Bengio. 2019.
\newblock Learning deep representations by mutual information estimation and
  maximization.
\newblock In \emph{Proc. of ICLR}.

\bibitem[{Hu and Liu(2004)}]{Hu2004}
Minqing Hu and Bing Liu. 2004.
\newblock Mining and summarizing customer reviews.
\newblock In \emph{Proc. of KDD}.

\bibitem[{Hyv{\"a}rinen and Oja(2000)}]{hyvarinen2000independent}
Aapo Hyv{\"a}rinen and Erkki Oja. 2000.
\newblock Independent component analysis: algorithms and applications.
\newblock \emph{Neural networks}, 13(4-5):411--430.

\bibitem[{Jernite et~al.(2017)Jernite, Bowman, and Sontag}]{Jernite17}
Yacine Jernite, Samuel~R. Bowman, and David~A. Sontag. 2017.
\newblock Discourse-based objectives for fast unsupervised sentence
  representation learning.
\newblock \emph{arXiv preprint arXiv:1705.00557}.

\bibitem[{Ji et~al.(2019)Ji, Henriques, and Vedaldi}]{ji2019}
Xu~Ji, F.~João Henriques, and Andrea Vedaldi. 2019.
\newblock Invariant information clustering for unsupervised image
  classification and segmentation.
\newblock \emph{arXiv preprint arXiv:1807.06653}.

\bibitem[{Kim(2014)}]{kim2014convolutional}
Yoon Kim. 2014.
\newblock Convolutional neural networks for sentence classification.
\newblock \emph{arXiv preprint arXiv:1408.5882}.

\bibitem[{Kiros et~al.(2015)Kiros, Zhu, Salakhutdinov, Zemel, Urtasun,
  Torralba, and Fidler}]{Kiros2015}
Ryan Kiros, Yukun Zhu, Russ~R Salakhutdinov, Richard Zemel, Raquel Urtasun,
  Antonio Torralba, and Sanja Fidler. 2015.
\newblock Skip-thought vectors.
\newblock In \emph{Proc. of NIPS}.

\bibitem[{Kong et~al.(2020)Kong, de~Masson~d'Autume, Yu, Ling, Dai, and
  Yogatama}]{kong2020}
Lingpeng Kong, Cyprien de~Masson~d'Autume, Lei Yu, Wang Ling, Zihang Dai, and
  Dani Yogatama. 2020.
\newblock A mutual information maximization perspective of language
  representation learning.
\newblock In \emph{Proc. of ICLR}.

\bibitem[{Le and Mikolov(2014)}]{Le2014}
Quoc~V. Le and Tomas Mikolov. 2014.
\newblock Distributed representations of sentences and documents.
\newblock In \emph{Proc. of ICML}.

\bibitem[{Li and Roth(2002)}]{li2002}
Xin Li and Dan Roth. 2002.
\newblock Learning question classifiers.
\newblock In \emph{Proc. of COLING}.

\bibitem[{Liu et~al.(2019)Liu, Ott, Goyal, Du, Joshi, Chen, Levy, Lewis,
  Zettlemoyer, and Stoyanov}]{liu2019roberta}
Yinhan Liu, Myle Ott, Naman Goyal, Jingfei Du, Mandar Joshi, Danqi Chen, Omer
  Levy, Mike Lewis, Luke Zettlemoyer, and Veselin Stoyanov. 2019.
\newblock Roberta: A robustly optimized bert pretraining approach.
\newblock \emph{arXiv preprint arXiv:1907.11692}.

\bibitem[{Logeswaran and Lee(2018)}]{logeswaran2018}
Lajanugen Logeswaran and Honglak Lee. 2018.
\newblock An efficient framework for learning sentence representations.
\newblock In \emph{Proc. of ICLR}.

\bibitem[{Marelli et~al.(2014)Marelli, Menini, Baroni, Bentivogli, Bernardi,
  and Zamparelli}]{marelli2014}
Marco Marelli, Stefano Menini, Marco Baroni, Luisa Bentivogli, Raffaella
  Bernardi, and Roberto Zamparelli. 2014.
\newblock A {SICK} cure for the evaluation of compositional distributional
  semantic models.
\newblock In \emph{Proc. of LREC}.

\bibitem[{Misra et~al.(2016)Misra, Ecker, and Walker}]{misra2016}
Amita Misra, Brian Ecker, and Marilyn Walker. 2016.
\newblock Measuring the similarity of sentential arguments in dialogue.
\newblock In \emph{Proc. of the 17th Annual Meeting of the Special Interest
  Group on Discourse and Dialogue}.

\bibitem[{Nowozin et~al.(2016)Nowozin, Cseke, and Tomioka}]{nowozin2016}
Sebastian Nowozin, Botond Cseke, and Ryota Tomioka. 2016.
\newblock f-gan: Training generative neural samplers using variational
  divergence minimization.
\newblock In \emph{Proc. of NIPS}.

\bibitem[{van~den Oord et~al.(2018)van~den Oord, Li, and Vinyals}]{oord2018}
A{\"{a}}ron van~den Oord, Yazhe Li, and Oriol Vinyals. 2018.
\newblock Representation learning with contrastive predictive coding.
\newblock \emph{arXiv preprint arXiv:1807.03748}.

\bibitem[{Pagliardini et~al.(2017)Pagliardini, Gupta, and
  Jaggi}]{pagliardini2017unsupervised}
Matteo Pagliardini, Prakhar Gupta, and Martin Jaggi. 2017.
\newblock Unsupervised learning of sentence embeddings using compositional
  n-gram features.
\newblock \emph{arXiv preprint arXiv:1703.02507}.

\bibitem[{Pang and Lee(2004)}]{pang2004}
Bo~Pang and Lillian Lee. 2004.
\newblock A sentimental education: Sentiment analysis using subjectivity
  summarization based on minimum cuts.
\newblock In \emph{Proc. of ACL}.

\bibitem[{Pang and Lee(2005)}]{pang2005}
Bo~Pang and Lillian Lee. 2005.
\newblock Seeing stars: Exploiting class relationships for sentiment
  categorization with respect to rating scales.
\newblock In \emph{Proc. of ACL}.

\bibitem[{Reimers et~al.(2016)Reimers, Beyer, and Gurevych}]{reimers2016}
Nils Reimers, Philip Beyer, and Iryna Gurevych. 2016.
\newblock Task-oriented intrinsic evaluation of semantic textual similarity.
\newblock In \emph{Proc. of COLING}.

\bibitem[{Reimers and Gurevych(2019)}]{reimers2019}
Nils Reimers and Iryna Gurevych. 2019.
\newblock Sentence-{BERT}: Sentence embeddings using {S}iamese {BERT}-networks.
\newblock In \emph{Proc. of EMNLP-IJCNLP}.

\bibitem[{Reimers et~al.(2019)Reimers, Schiller, Beck, Daxenberger, Stab, and
  Gurevych}]{reimers2019b}
Nils Reimers, Benjamin Schiller, Tilman Beck, Johannes Daxenberger, Christian
  Stab, and Iryna Gurevych. 2019.
\newblock Classification and clustering of arguments with contextualized word
  embeddings.
\newblock In \emph{Proc. of ACL}.

\bibitem[{Socher et~al.(2011)Socher, Huang, Pennin, Manning, and
  Ng}]{Socher2011}
Richard Socher, Eric~H. Huang, Jeffrey Pennin, Christopher~D Manning, and
  Andrew~Y. Ng. 2011.
\newblock Dynamic pooling and unfolding recursive autoencoders for paraphrase
  detection.
\newblock In \emph{Proc. of NIPS}.

\bibitem[{Socher et~al.(2013)Socher, Perelygin, Wu, Chuang, Manning, Ng, and
  Potts}]{socher2013}
Richard Socher, Alex Perelygin, Jean Wu, Jason Chuang, Christopher~D. Manning,
  Andrew Ng, and Christopher Potts. 2013.
\newblock Recursive deep models for semantic compositionality over a sentiment
  treebank.
\newblock In \emph{Proc. of EMNLP}.

\bibitem[{Sun et~al.(2020)Sun, Hoffmann, and Tang}]{sun2019infograph}
Fan-Yun Sun, Jordan Hoffmann, and Jian Tang. 2020.
\newblock Infograph: Unsupervised and semi-supervised graph-level
  representation learning via mutual information maximization.
\newblock In \emph{Proc. of ICLR}.

\bibitem[{Velickovic et~al.(2019)Velickovic, Fedus, Hamilton, Li{\`o}, Bengio,
  and Hjelm}]{Velickovic2018DeepGI}
Petar Velickovic, William Fedus, William~L. Hamilton, Pietro Li{\`o}, Yoshua
  Bengio, and R.~Devon Hjelm. 2019.
\newblock Deep graph infomax.
\newblock In \emph{Proc. of ICLR}.

\bibitem[{Wiebe et~al.(2005)Wiebe, Wilson, and Cardie}]{Wiebe2005}
Janyce Wiebe, Theresa Wilson, and Claire Cardie. 2005.
\newblock Annotating expressions of opinions and emotions in language.
\newblock \emph{Language Resources and Evaluation}.

\bibitem[{Williams et~al.(2018)Williams, Nangia, and Bowman}]{williams2018}
Adina Williams, Nikita Nangia, and Samuel Bowman. 2018.
\newblock A broad-coverage challenge corpus for sentence understanding through
  inference.
\newblock In \emph{Proc. of NAACL-HLT}.

\bibitem[{Yang et~al.(2018)Yang, Yuan, Cer, yi~Kong, Constant, Pilar, Ge, Sung,
  Strope, and Kurzweil}]{Yang2018LearningST}
Yinfei Yang, Steve Yuan, Daniel~Matthew Cer, Sheng yi~Kong, Noah Constant,
  P.~Pilar, Heming Ge, Yun-Hsuan Sung, B.~Strope, and R.~Kurzweil. 2018.
\newblock Learning semantic textual similarity from conversations.
\newblock In \emph{Proc. of RepL4NLP@ACL}.

\bibitem[{Zhu et~al.(2015)Zhu, Kiros, Zemel, Salakhutdinov, Urtasun, Torralba,
  and Fidler}]{zhu2015}
Yukun Zhu, Ryan Kiros, S.~Richard Zemel, Ruslan Salakhutdinov, Raquel Urtasun,
  Antonio Torralba, and Sanja Fidler. 2015.
\newblock Aligning books and movies: Towards story-like visual explanations by
  watching movies and reading books.
\newblock In \emph{Proc. of ICCV}.

\end{thebibliography}
\bibliographystyle{acl_natbib}

\end{document}